\documentclass[journal]{IEEEtran}

\usepackage{bm}

\usepackage{booktabs}
\usepackage{float}

\usepackage{amsmath,amsfonts}
\usepackage[noend]{algpseudocode}
\usepackage{algorithm}
\usepackage{array}
\usepackage{multirow}
\usepackage{booktabs}
\usepackage{pifont}
\usepackage[flushleft]{threeparttable}
\usepackage{dsfont}
\usepackage{textcomp}
\usepackage{stfloats}
\usepackage{url}
\usepackage{verbatim}
\usepackage{graphicx}
\usepackage{cite}
\usepackage{graphicx}
\usepackage{lipsum}

\usepackage{placeins}

\usepackage[table]{xcolor}
\usepackage{rotating}
\usepackage{lipsum}
\usepackage{eqparbox}
\usepackage{etoolbox}  
\usepackage[acronym]{glossaries}
\usepackage{colortbl} 
\usepackage{color, colortbl}

\usepackage[hidelinks,colorlinks=true,linkcolor=blue,urlcolor=blue,citecolor=blue]{hyperref}

\usepackage{setspace} 

\usepackage{makecell}

\usepackage{collcell}
\usepackage{pgf}
\usepackage{pgfplots}
\pgfplotsset{compat=1.18}
\usepackage{pgfplotstable}
\usepackage{hhline}
\usepackage{multirow}

\usepackage{tikz}
\usepackage[margin=1in]{geometry}
\usetikzlibrary{positioning, shapes.multipart, arrows.meta}

\def\colorModel{hsb} 

\newcommand\ColCell[1]{
  \pgfmathparse{#1<50?1:0}  
    \ifnum\pgfmathresult=0\relax\color{white}\fi
  \pgfmathsetmacro\compA{0}      
  \pgfmathsetmacro\compB{#1/100} 
  \pgfmathsetmacro\compC{1}      
  \edef\x{\noexpand\centering\noexpand\cellcolor[\colorModel]{\compA,\compB,\compC}}\x #1
  } 
\newcolumntype{E}{>{\collectcell\ColCell}m{0.4cm}<{\endcollectcell}}  

\algnewcommand\algorithmicforeach{\textbf{for each}}
\algdef{S}[FOR]{ForEach}[1]{\algorithmicforeach\ #1\ \algorithmicdo}

\definecolor{LightCyan}{rgb}{0.88,1,1}
\usepackage{todonotes}
\usepackage[hidelinks]{hyperref}
\usepackage[capitalise]{cleveref} 
\begin{document}

\title{Physics-Informed Machine Learning for Transformer Condition Monitoring -- Part II: Physics-Informed Neural Networks and Uncertainty Quantification}

\author{Jose I. Aizpurua,~\IEEEmembership{Senior Member,~IEEE}

\thanks{J. Aizpurua is with the University of the Basque Country (UPV/EHU), Faculty of Informatics, Department of Computer Science and Artificial Intelligence, Donostia - San Sebastian, Spain.}
\thanks{978-84-XX-XXXXX-X /25 © 2025 Red iNtransf}
\thanks{\textcolor{red}{“© 2025 IEEE.  Personal use of this material is permitted.  Permission from IEEE must be obtained for all other uses, in any current or future media, including reprinting/republishing this material for advertising or promotional purposes, creating new collective works, for resale or redistribution to servers or lists, or reuse of any copyrighted component of this work in other works.”}}
}

\markboth{2025 8th International Advanced Research Workshop on Transformers (ARWtr)  –   Baiona - Spain, (12)13-15 October 2025}%
{Shell \MakeLowercase{\textit{et al.}}: Bare Demo of IEEEtran.cls for IEEE Journals}

\maketitle

\begin{abstract} 
The integration of physics-based knowledge with machine learning models is increasingly shaping the monitoring, diagnostics, and prognostics of electrical transformers. In this two-part series, the first paper introduced the foundations of Neural Networks (NNs) and their variants for health assessment tasks. This second paper focuses on integrating physics and uncertainty into the learning process. We begin with the fundamentals of Physics-Informed Neural Networks (PINNs), applied to spatiotemporal thermal modeling and solid insulation ageing. Building on this, we present Bayesian PINNs as a principled framework to quantify epistemic uncertainty and deliver robust predictions under sparse data. Finally, we outline emerging research directions that highlight the potential of physics-aware and trustworthy machine learning for critical power assets.
\end{abstract}



\section{Introduction}
\label{sec:Intro}

\IEEEPARstart{P}{hysics}-informed Neural Networks (PINNs) have emerged as a promising approach to address some of the limitations of classical Neural Networks (NNs) in science and engineering applications. In particular, two major challenges are often cited: (i) the absence of an explicit physical reasoning mechanism and (ii) the difficulty of operating reliably under scarce data conditions. Building on the foundations of Neural Networks and Reinforcement Learning introduced in Part I, this second paper focuses on PINNs for transformer applications, and extends them to uncertainty quantification via Bayesian Neural Networks (BNNs).

PINNs enhance conventional NNs by embedding physics-based models directly into the learning process \cite{Ramirez_25}. Unlike traditional hybrid approaches that integrate physics-based and data-driven models in series or parallel configurations \cite{BPINN_PHM}, such as error-correction models for oil temperature estimation \cite{Aizpurua23}, PINNs integrate physics and data simultaneously during training. This leads to models that not only capture empirical patterns, but also enforce consistency with governing equations, thereby improving generalization while reducing dependence on large labeled datasets.

In the context of Prognostics and Health Management (PHM), and particularly for prognostics models that aim to predict the Remaining Useful Life (RUL) of an asset, uncertainty quantification (UQ) is critical for reliable future ageing predictions \cite{Sankaraman15}. In the UQ literature, uncertainty is commonly divided into two categories: aleatoric uncertainty, which arises from noisy or incomplete data (e.g., sensor errors), and epistemic uncertainty, which reflects the limitations of the model or incomplete knowledge (e.g., network architecture or parameter uncertainty). In this work, we focus on epistemic uncertainty, which directly affects the ability of NNs to learn and generalize.

The remainder of this paper is structured as follows. Section~\ref{Sec:PINNs} introduces PINNs for embedding physics into NN training. Section~\ref{s:BPINNs} extends PINNs to Bayesian PINNs for epistemic uncertainty quantification. Finally, Section~\ref{Sec:Conclusions} summarizes the findings and outlines future research directions in machine learning for transformer health monitoring.

\section{Physics-Informed Neural Networks}
\label{Sec:PINNs}

\subsection{Motivation: Neural Networks Constrained by Physics}

Classical Neural Networks and their variants, such as Convolutional NNs (CNNs), excel in pattern recognition tasks when abundant labeled data is available. However, in many industrial applications, including transformer condition monitoring, labeled data is scarce, expensive to obtain, and often does not cover the full range of operating conditions.

At the same time, the behaviour of these systems is governed by well-understood physical laws, such as heat transfer or fluid dynamics. PINNs provide a framework to incorporate such knowledge directly into the Neural Network training process \cite{Raissi_19}. By constraining the NN with governing equations, PINNs learn from both sparse data and encoded physics, ensuring physically consistent outputs and improved extrapolation.

\noindent Key motivations for using PINNs include:
\begin{itemize}
    \item Physical consistency. Ensures predictions comply with encoded physical laws.
    \item Improved generalization. Enables extrapolation beyond the training regime.
    \item Interpretability. Outputs remain linked to measurable physical quantities.
    \item Data efficiency. Physics constraints act as implicit regularizers.
\end{itemize}

\subsection{Core Concepts}

A PINN models the solution of a physical system, typically described by a Partial Differential Equation (PDE), using a Neural Network $u_\theta(x,t)$ with trainable parameters $\bm{\theta}$. The PDE can be written generically as \cite{Raissi_19}:

\begin{equation}
u_t+\mathcal{N}[u;\lambda] = 0, \quad  x \in \Omega, \quad t \in [0,T],
\end{equation}

\noindent where $u(t,x)$ denotes the solution, $\mathcal{N}[\cdot;\lambda]$ is a nonlinear differential operator parameterized by $\lambda$, and $\Omega$ is the spatiotemporal domain.

For fixed model parameters $\lambda$, the goal is to solve $u_t+\mathcal{N}[u] = 0, x \in \Omega, t \in [0,T]$, by defining the residual function, $r(t,x)$:

\begin{equation}
r(t,x):=u_t+\mathcal{N}[u].
\end{equation}

\noindent The PINN seeks an NN approximation $u_\theta(x,t)$ with $\bm{\theta}=\{w,b\}$ being the set of NN's weight and bias parameters, that minimizes the residual across the domain while also satisfying boundary and initial conditions. This is achieved through a composite loss function $\mathcal{L}(\bm{\theta})$:

\begin{equation}
\label{eq:loss_pinn}
\mathcal{L}(\bm{\theta}) = \lambda_{\text{0}} \cdot \mathcal{L}_{\text{0}}(\bm{\theta}) + \lambda_{\text{PDE}} \cdot \mathcal{L}_{\text{r}}(\bm{\theta}) + \lambda_{\text{BC}} \cdot \mathcal{L}_{\text{BC}}(\bm{\theta}),
\end{equation}

\noindent with,

\begin{equation}
    \mathcal{L}_{\text{0}}(\bm{\theta}) = \frac{1}{N_0}\sum_{i=1}^{N_0} |\hat{u}(x_i,0)-u(x_i,0)|^2
\end{equation}

\begin{equation}
    \mathcal{L}_{\text{r}}(\bm{\theta}) = \sum_{i=1}^{N_r} |r(x_i,t_i)|^2
\end{equation}

\begin{equation}
    \mathcal{L}_{\text{BC}}(\bm{\theta}) = \frac{1}{N_{BC}}\sum_{i=1}^{N_{BC}} |\hat{u}(x_i,0)-u(x_i,t_i)|^2
\end{equation}

\noindent where $\bm{\theta}=\{w,b\}$, $N_0$, $N_{BC}$, and $N_r$ are the number of initial conditions, boundary conditions and residual points respectively, $u(x_i,t_i)$ and $r(x_i,t_i)$ denote the known solution and the residual of the PDE, respectively, for each training point $i$ defined at coordinates $(x_i,t_i)$, and $\lambda_0$, $\lambda_{BC}$, and $\lambda_r$ are the global multipliers for initial conditions, boundary conditions, and residual terms, respectively.

The specification of local multipliers to normalize the loss function is a challenging activity, refer to \cite{Ramirez_25} for potential solutions. Moreover, it may be the case that the loss function in Eq.~(\ref{eq:loss_pinn}) includes an additional data loss term, including additional measurements that can act as regularization terms during the PINN training process. 

\begin{figure}[!htb]
\centering
\includegraphics[width=0.95\columnwidth]{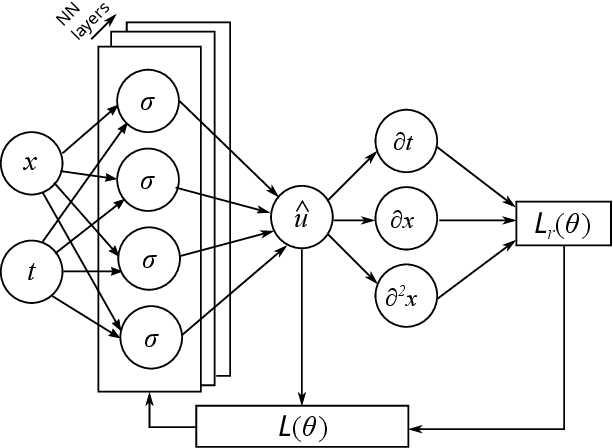}
\caption{Generic PINN structure with spatio-temporal coordinates as inputs $x,t$, predictions $\hat{u}$ at the output, temporal and spatial derivatives that define the residual loss $L_r(\bm{\theta})$, and the total loss $L(\bm{\theta})$ with respect to the NN parameters $\bm{\theta}$.}
\label{fig:PINN_structure}
\end{figure}

Automatic differentiation enables efficient computation of PDE residuals with respect to NN outputs, $\hat{u}$, ensuring that the governing equations are enforced throughout the domain, not just at measurement points. Fig.~\ref{fig:PINN_structure} illustrates the generic PINN architecture.

Using the composite loss (\ref{eq:loss_pinn}), the network parameters $\bm{\theta}$ are updated via backpropagation and gradient-based optimization.

\subsection{Case Study: Transformer Thermal Ageing}

Transformer insulation ageing is mainly driven by temperature-dependent chemical processes. Therefore, accurately estimating the internal temperature distribution is crucial for predicting insulation health. In practice, however, internal temperature sensors are sparse and invasive, making direct monitoring difficult.

A simplified yet effective model is the one-dimensional (1D) heat-diffusion equation:

\small
\begin{equation}
\label{eq:PDE_1D_Diffusion}
\begin{split}
 k\frac{\partial^2\Theta_{O}(x,t)}{\partial x^2} + q(x,t) &= \rho c_p \frac{\partial \Theta_{O}(x,t)}{\partial t},
\end{split}
\end{equation}
\normalsize

\noindent where:
\begin{itemize}
    \item $x, t \in \mathbb{R}$ are spatial and temporal coordinates,
    \item $\Theta_{O}(x,t)$ is the oil temperature [K],
    \item $k$ is thermal conductivity [W/m·K],
    \item $\rho$ is oil density [kg/m\textsuperscript{3}],
    \item $c_p$ is specific heat capacity [J/kg·K],
    \item $q(x,t)$ is internal heat generation [W/m\textsuperscript{3}],
\end{itemize}

Defining the thermal diffusivity $\alpha = \tfrac{k}{\rho c_p}$ [m\textsuperscript{2}/s], the PDE can be rewritten as:

\begin{equation}
\frac{\partial^2 \Theta_{O}(x,t)}{\partial x^2} + \frac{1}{k} q(x,t) = \frac{1}{\alpha}\frac{\partial \Theta_{O}(x,t)}{\partial t}.
\end{equation}

The heat source term $q(x,t)$ accounts for load and no-load losses, as well as convective cooling:

\begin{equation}
\label{eq:heat-source}
    q(x,t) = P_0 + P_K(t) - h \left(\Theta_{O}(x,t) - \Theta_A(t) \right),
\end{equation}

\noindent where:
\begin{itemize}
    \item $P_0$ is no-load loss [W],
    \item $P_K(t) = K(t)^2 \mu$ is load loss, where $K(t)$ is load factor [p.u.] and $\mu$ is the rated loss [W],
    \item $h$ is convective heat-transfer coefficient [W/m$^2\cdot$K],
    \item $\Theta_A(t)$ is ambient temperature [K].
\end{itemize}

Boundary conditions are imposed using accessible measurements:

\begin{equation}
\label{eq:BoundaryConditions}
\begin{split}
    \Theta_{O}(0,t) &= \Theta_{A}(t), \\
    \Theta_{O}(H,t) &= \Theta_{TO}(t),
\end{split}
\end{equation}

\noindent where $H$ is the height of the transformer tank and $\Theta_{TO}(t)$ is the measured top-oil temperature.

\begin{figure}[!htb]
	\centering
	\includegraphics[width=\columnwidth]{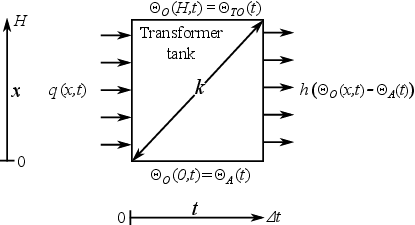}
	\caption{Schematic of the transformer heat diffusion model, including heat sources and boundary conditions.}
	\label{fig:Trafo_Diffusion_Basic}
\end{figure}

Fig.~\ref{fig:Trafo_Diffusion_Basic} shows the thermal parameters of the transformer of the heat diffusion model  \cite{Ramirez_25}, which considers the heat source, $q(x,t)$, and the convective heat transfer, $h(\Theta_{O}(x,t)-\Theta_{A}(t))$.

\begin{figure}[!htb]
	\centering
        \includegraphics[width=\columnwidth]{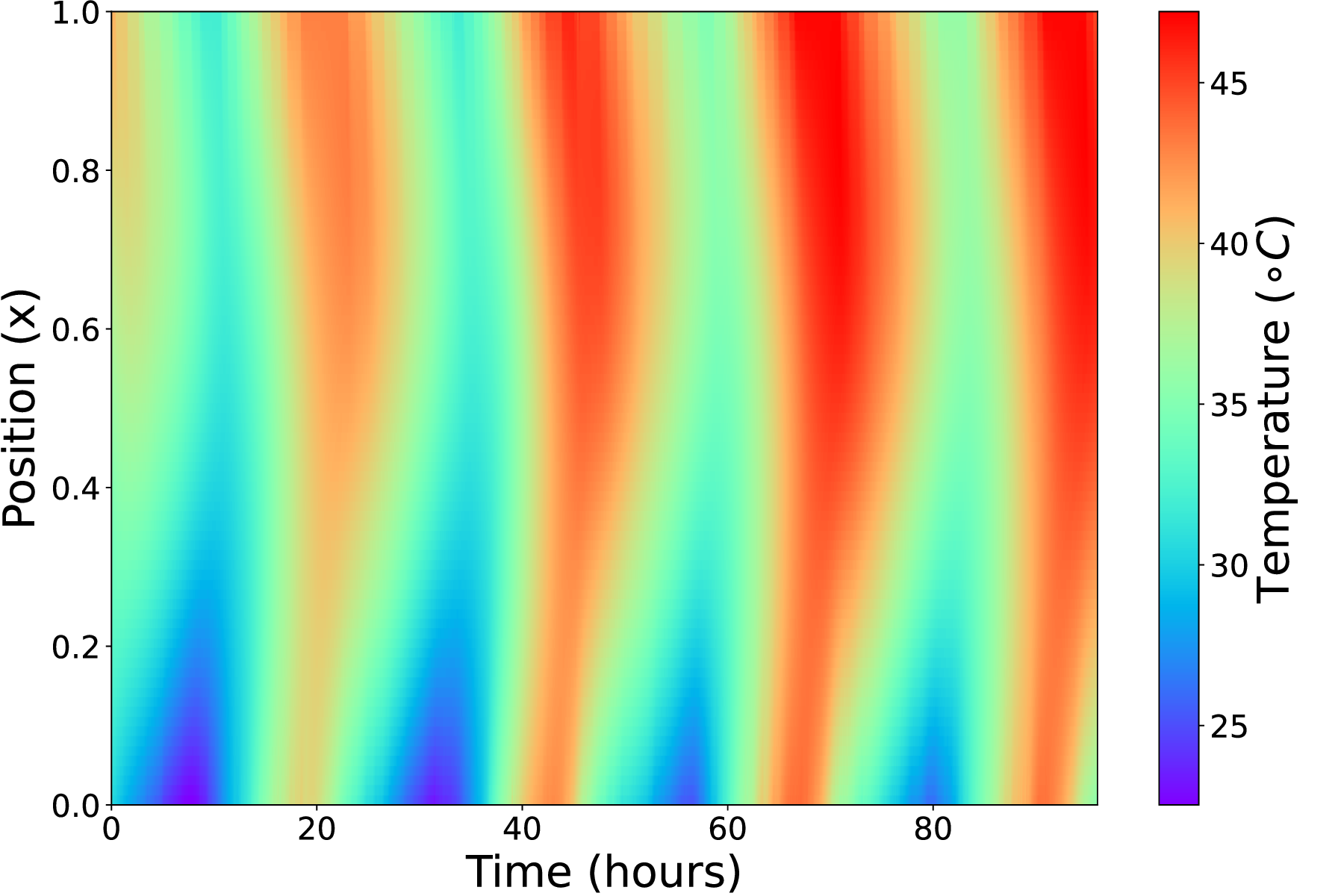}
	\caption{Reference temperature distribution obtained via FEM simulation.}
	\label{fig:PDEmatlab}
\end{figure}

Given the input profiles $\{K(t), \Theta_A(t), \Theta_{TO}(t)\}$, the PDE in Eq.~(\ref{eq:PDE_1D_Diffusion}) can be solved numerically, e.g., with Finite Element Methods (FEM). Fig.~\ref{fig:PDEmatlab} shows the FEM reference solution for a representative case study \cite{Ramirez_25}.

\begin{figure}[!htb]
	\centering
        \includegraphics[width=\columnwidth]{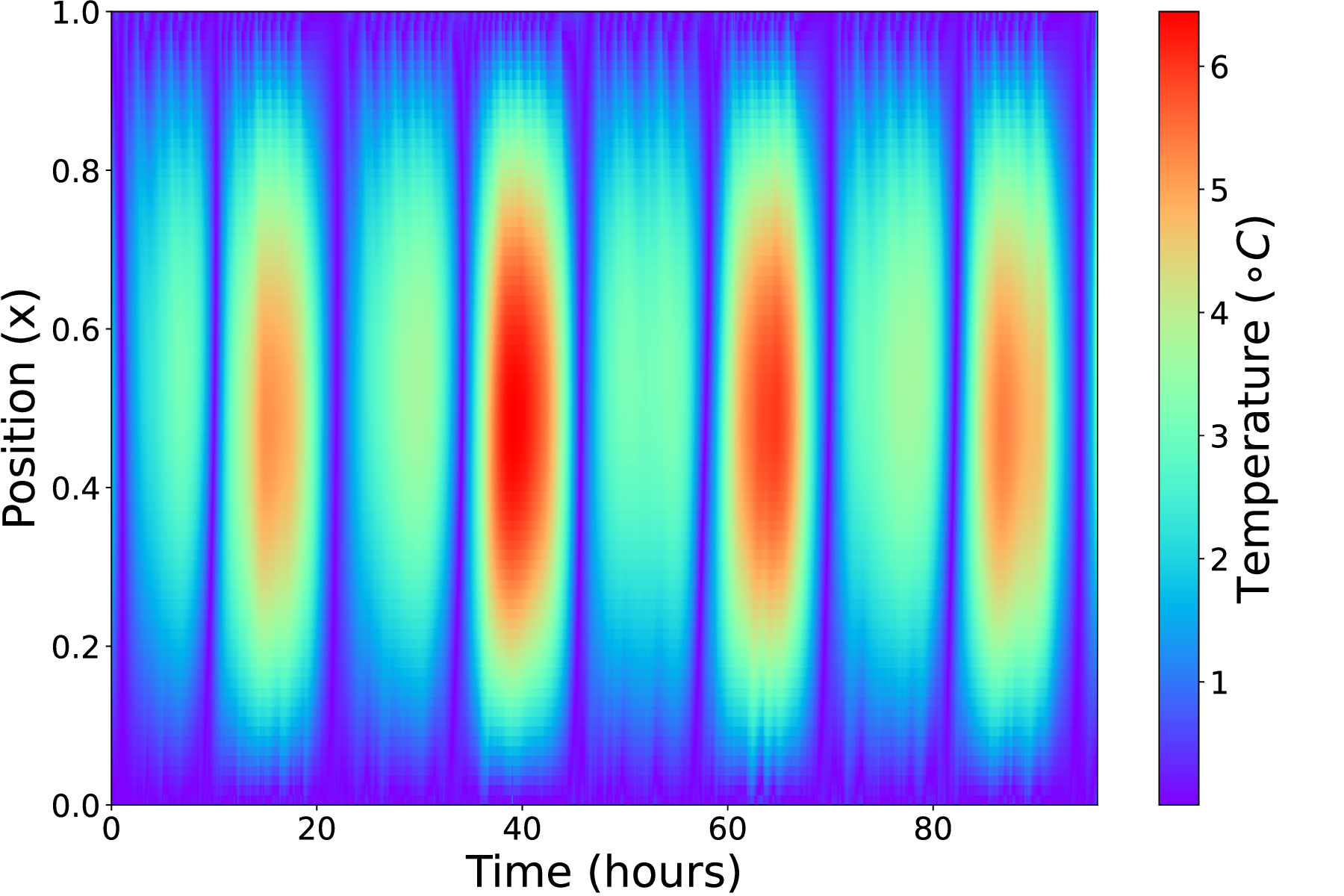}
	\caption{Prediction error of the PINN model compared to FEM.}
	\label{fig:Error_PINN}
\end{figure}

Instead of solving Eq.~(\ref{eq:PDE_1D_Diffusion}) directly, a PINN can be trained to approximate $\Theta_O(x,t)$ across the spatiotemporal domain. Using the loss function in Eq.~(\ref{eq:loss_pinn}), the NN learns not only from available measurements, but also from the PDE residuals and boundary conditions. Fig.~\ref{fig:Error_PINN} shows the prediction error of the PINN compared to the FEM baseline.

The results confirm that the PINN achieves accurate temperature estimation, closely following the FEM solution. Compared with numerical solvers, the PINN requires significantly less computational effort once trained. Moreover, PINNs generalize to unseen loading and ambient conditions, leveraging physics to avoid overfitting. The simplest vanilla PINN model can be further improved through different mechanisms. Refer to \cite{Ramirez_25} for a complete set of solutions.

\begin{figure}[!htb]
	\centering
	\includegraphics[width=\columnwidth]{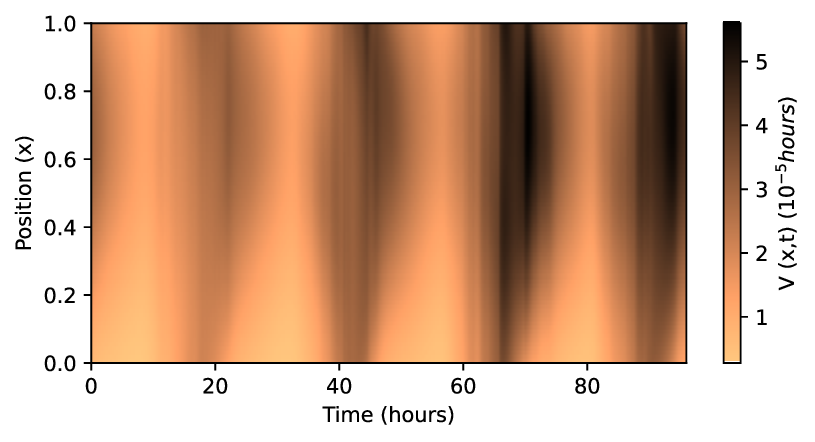}
        \caption{Instantaneous transformer insulation spatial ageing, $V(x,t)$.}
	\label{fig:Spatial_Ageing_Faa}
\end{figure}

Importantly, PINNs enable the extension of classical health indicators. For instance, insulation ageing is often estimated from bulk temperature values \cite{Aizpurua23,Aizpurua_TII,Aizpurua_22}. By incorporating spatiotemporal estimates (Eq.~(\ref{eq:PDE_1D_Diffusion})), PINNs allow for spatially resolved ageing assessment. Fig.~\ref{fig:Spatial_Ageing_Faa} illustrates the instantaneous spatial ageing distribution derived from Arrhenius’ law applied to the PINN-predicted temperature field.

This case study highlights the potential of PINNs to enable data-efficient, physics-consistent transformer health monitoring, paving the way for advanced digital twins that integrate thermal and ageing dynamics.

\section{Bayesian Physics-informed Neural Networks}
\label{s:BPINNs}

\subsection{Motivation}

While PINNs successfully embed physical laws into Neural Network training, they provide deterministic predictions. For critical applications such as transformer health monitoring, deterministic outputs are not sufficient and asset management teams and operators also need to understand the confidence and reliability of model predictions. This is especially important under sparse data and uncertain future load/ambient conditions.

Uncertainty quantification (UQ) is generally divided into \cite{Psaros23}:

\begin{itemize}
    \item Aleatoric uncertainty: inherent variability from noisy measurements or stochastic phenomena.
    \item Epistemic uncertainty: uncertainty due to incomplete knowledge, e.g., limited data or model misspecification.
\end{itemize}

In this work, we focus on epistemic uncertainty, which directly reflects how well the Neural Network generalizes under limited data. To capture it, we extend PINNs into a Bayesian framework, resulting in Bayesian PINNs (B-PINNs).

\subsection{Core Concepts}

In standard NNs, weights $\bm{\theta}$ are treated as deterministic parameters. In Bayesian Neural Networks (BNNs), weights are instead modeled as probability distributions \cite{BNN_tutorial}. Given data $\mathcal{D}=\{x^{(i)},y^{(i)}\}$, Bayesian inference updates prior beliefs $p(\bm{\theta})$ into a posterior distribution:

\begin{equation}
\label{eq:Bayes}
p(\bm{\theta}|\mathcal{D}) = \frac{p(\mathcal{D}|\bm{\theta}) p(\bm{\theta})}{p(\mathcal{D})},
\end{equation}

\noindent where $p(\mathcal{D}|\bm{\theta})$ is the likelihood, $p(\bm{\theta})$ the prior, and $p(\mathcal{D})$ the marginal likelihood.

The likelihood is often estimated the individual product of pointwise estimated likelihoods $p_i(\mathcal{D}|\bm{\theta})$ based on the normal distribution $N(\mu,\sigma)$ defined as follows:

\begin{equation}
	P(\mathcal{D}|\bm{\theta})=\prod_{i=0}^{N}  p(y^{(i)}|x^{(i)},\bm{\theta})
\end{equation}

\begin{equation}
	\label{eq:Gaussian}
	p(y^{(i)}|x^{(i)},\bm{\theta})=\frac{1}{\sqrt{2\pi}\sigma}exp(-\frac{||x_i-x(t_i)||^2}{2\sigma^2})
\end{equation}

For the sake of simplicity we will consider a Gaussian prior (with mean zero and unit variance). See \cite{BPINN_PHM} and references herein for other alternatives.

The analytical posterior in BNNs, $p(\bm{\theta}|\mathcal{D})$, is generally intractable, and therefore approximation methods are required. In this work, variational inference (VI) is used to approximate the true posterior $p(\bm{\theta}|\mathcal{D})$ with a tractable variational distribution $q(\bm{\theta}|\bm{\phi})$ of known functional form, parameterized by $\bm{\phi}$, by minimizing the Kullback–Leibler (KL) divergence between them.

The approximation is optimized by maximizing the Evidence Lower Bound (ELBO), which balances (i) a KL divergence term between the variational distribution and the prior $p(\bm{\theta})$ (known as model complexity), and (ii) a data fit term (likelihood). The corresponding optimization objective, can be approximated by Monte Carlo sampling, drawing $N$ samples $\bm{\theta}^{(i)}$ from $q(\bm{\theta}|\bm{\phi})$ \cite{BNN_tutorial}:

\small
\begin{equation}
	\mathcal{L}(\mathcal{D},\!\bm{\phi})\! \approx \!
	\frac{1}{N}\! \sum_{i=1}^{N}\!\big[
	\log q(\bm{\theta}^{(i)}|\bm{\phi}) 
	\!-\! \log p(\bm{\theta}^{(i)}) 
	\!-\! \log p(\mathcal{D}|\bm{\theta}^{(i)}) \big]
	\label{eq:MC_ELBO}
\end{equation}
\normalsize

Throughout this paper, the variational posterior is assumed to follow a Gaussian distribution $\bm{\phi}=(\bm{\mu},\bm{\sigma})$, where $\bm{\mu}$ is the mean vector of the distribution and $\bm{\sigma}$ is the standard deviation vector.

\subsection{B-PINN Formulation}

Recall that PINNs define a composite loss (Eq.~(\ref{eq:loss_pinn})) that enforces data, PDE residuals, and boundary/initial conditions. In a Bayesian setting, these terms enter through the likelihood, $P(\mathcal{D}|\Theta)$, which is estimated assuming a Gaussian distribution (cf. Eq.~(\ref{eq:Gaussian})), defined as follows:

\footnotesize
\begin{align}
	\begin{split}
		& p(u_0 | x_0, \bm{\theta}^{(i)}) = \prod_{i=1}^{N_0} \frac{1}{\sqrt{2\pi} \sigma_0}\exp\left(\!-\frac{\| \hat{u}(x_0, 0; \bm{\theta}^{(i)}) - u_0 \|^2}{2 \sigma_0^2}\right)\\
		& p(\!u_{bc} | x_{bc}, t_{bc}, \bm{\theta}^{(i)}\!) \!=\! \prod_{i=1}^{N_{bc}}\! \frac{1}{\sqrt{2\pi} \sigma_{bc}}\!\exp\!\left(\!-\frac{\| \!\hat{u}(\!x_{bc}, t_{bc}; \bm{\theta}^{(i)}\!) \!- \!u_{bc}\! \|^2}{2 \sigma_{bc}^2}\!\right)\\
		& p(r(x_f, t_f; \bm{\theta}^{(i)})) = \prod_{i=1}^{N_{f}} \frac{1}{\sqrt{2\pi} \sigma_{f}}\exp\left(\!-\frac{\| r(x_f, t_f; \bm{\theta}^{(i)}) \|^2}{2 \sigma_{f}^2}\right)
	\end{split}
	\label{eq:L_BPINN}
\end{align}
\normalsize

\noindent where $\sigma_0$, $\sigma_{bc}$, and $\sigma_f$ denote the standard deviation of initial, boundary, and residual points.

\begin{figure*}[!htb]
	\centering
	\includegraphics[width=2\columnwidth]{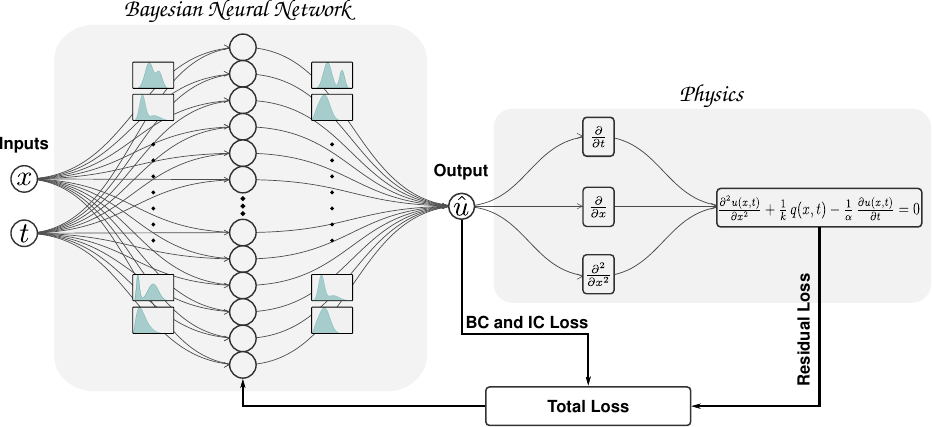}
	\caption{Proposed Bayesian-PINN approach for the probabilistic spatiotemporal transformer thermal model.}
	\label{fig:BPINN_Framework_General}
\end{figure*}

For computational tractability and efficiency, the log-likelihood terms are considered by taking the logarithm of Eq. (\ref{eq:L_BPINN}). Subsequently, the ELBO loss defined for a generic BNN [cf. Eq.~(\ref{eq:MC_ELBO})], is adapted for B-PINN posterior inference:

\begin{align}
\begin{split}
		\mathcal{L}^{(i)} &= 
		\log q(\bm{\theta}^{(i)}|\bm{w}) 
		- \log p(\bm{\theta}^{(i)}) \\ 
		&- \lambda_0\log p(u_0 | x_0, \bm{\theta}^{(i)}) \\
		&- \lambda_b\log p(u_{bc} | x_{bc}, t_{bc}, \bm{\theta}^{(i)}) \\
		&-\lambda_r \log p(r | x_f, t_f, \bm{\theta}^{(i)})
	\label{eq:ELBO_BPINN}
\end{split}
\end{align}

This process results in the approximation of the variational posterior distribution  $q(\bm{\theta}|\bm{w})$.

This process results in the approximation of the variational posterior distribution  $q(\bm{\theta}|\bm{w})$. The training process of the B-PINN approach is summarized in Algorithm~\ref{alg:Bayesian_PINN}.

\begin{algorithm}[!htb]
	\caption{B-PINN Training via Variational Inference}
	\label{alg:Bayesian_PINN}
	\begin{algorithmic}[1]
		\State \textbf{Input:} Collocation points $(x_f, t_f)$, initial condition data $(x_0, u_0)$, boundary condition data $(x_{bc}, t_{bc}, u_{bc})$, prior distribution $p(\bm{\theta})$
		\State Initialize variational parameters $\bm{w} = \{\bm{\mu}, \bm{\sigma}\}$ for $\bm{\theta}$
		\While{not converged}
		\State Sample $\bm{\theta}^{(i)} \sim q(\bm{\theta}|\bm{w})$ via reparameterization trick
		\State Compute outputs: $\hat{u}(x,t;\bm{\theta}^{(i)})$
		\State Compute PDE residual: $r(x_f,t_f;\bm{\theta}^{(i)})$
		\State Evaluate log-likelihood terms taking the $\log$ of Eq. (\ref{eq:L_BPINN})
		\State Evaluate prior log-probability: $\log p(\bm{\theta}^{(i)})$
		\State Evaluate variational density: $\log q(\bm{\theta}^{(i)}|\bm{w})$
		\State Compute Monte Carlo estimate of ELBO loss via Eq. (\ref{eq:ELBO_BPINN})
		\State Update $\bm{w} = \{\bm{\mu}, \bm{\sigma}\}$ using gradients $\nabla_{\bm{w}} \mathcal{L}^{(i)}$
		\EndWhile
		\State \textbf{Return:} Variational posterior $q(\bm{\theta}|\bm{w})$
	\end{algorithmic}
\end{algorithm}

\subsection{Case Study: Probabilistic Transformer Thermal Modelling}

We apply the B-PINN framework to the same 1D heat diffusion problem introduced in Section~\ref{Sec:PINNs}. Fig.~\ref{fig:BPINN_Framework_General} summarizes the architecture.

\begin{figure}[!htb]
	\centering
	\includegraphics[width=\columnwidth]{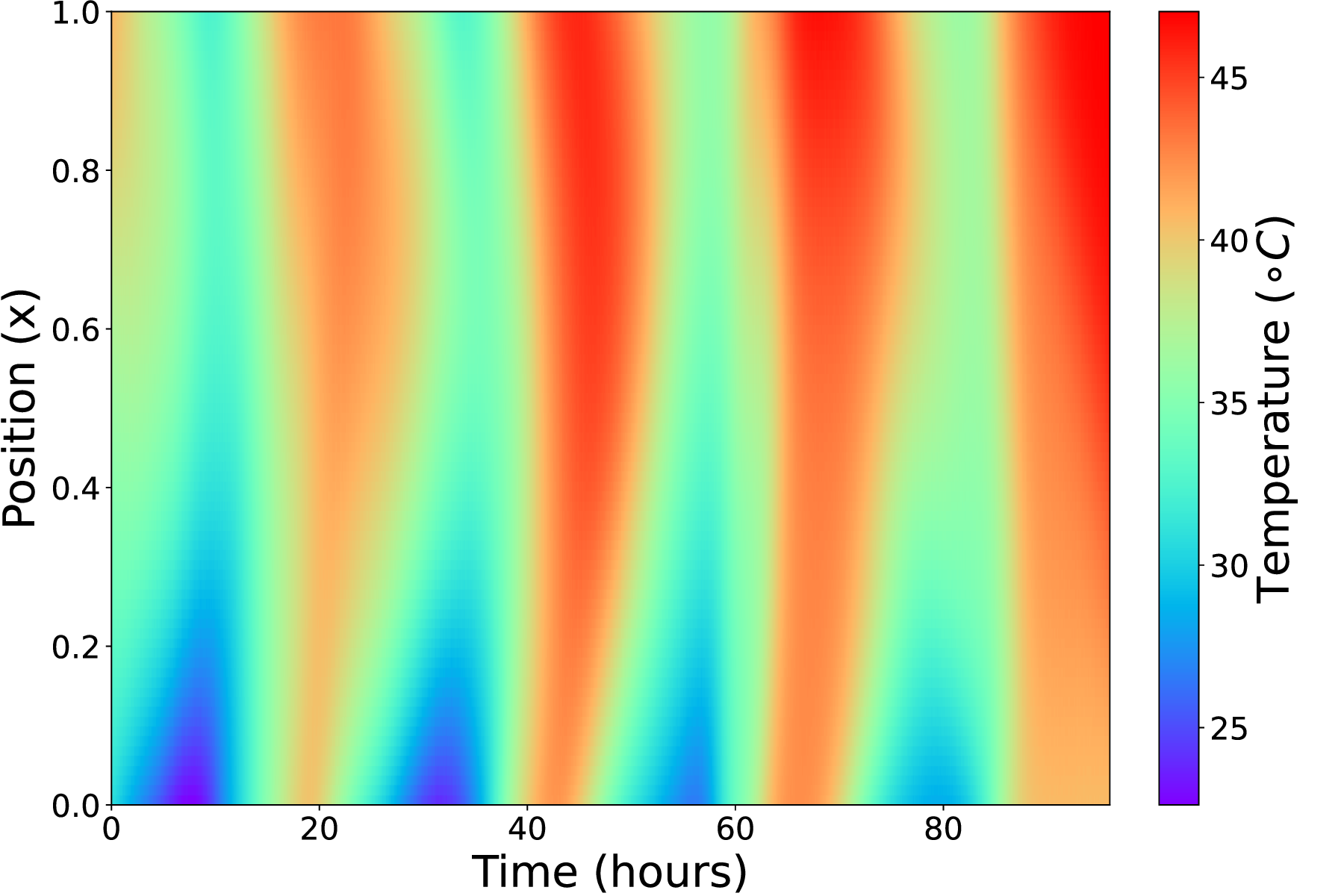}
	\includegraphics[width=\columnwidth]{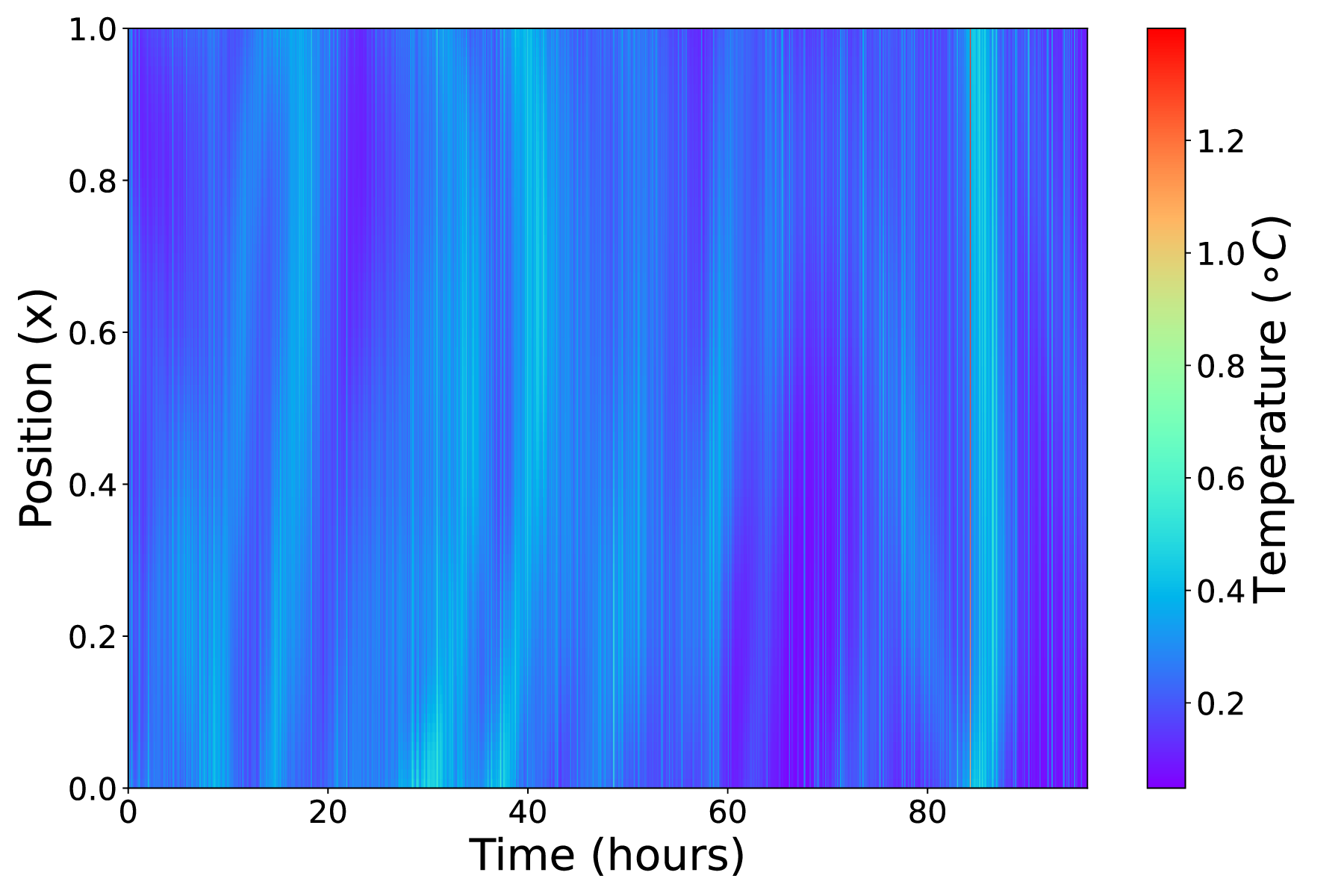}
	\caption{B-PINN posterior predictions: (a) mean oil temperature; (b) associated uncertainty (standard deviation).}
	\label{fig:MeanTemperatureEstimation_BPINN}
\end{figure}

Fig.~\ref{fig:MeanTemperatureEstimation_BPINN} shows the posterior mean and standard deviation of the oil temperature distribution. The mean closely matches FEM reference results, while the standard deviation quantifies predictive uncertainty.

\begin{figure}[!htb]
	\centering
	\includegraphics[width=\columnwidth]{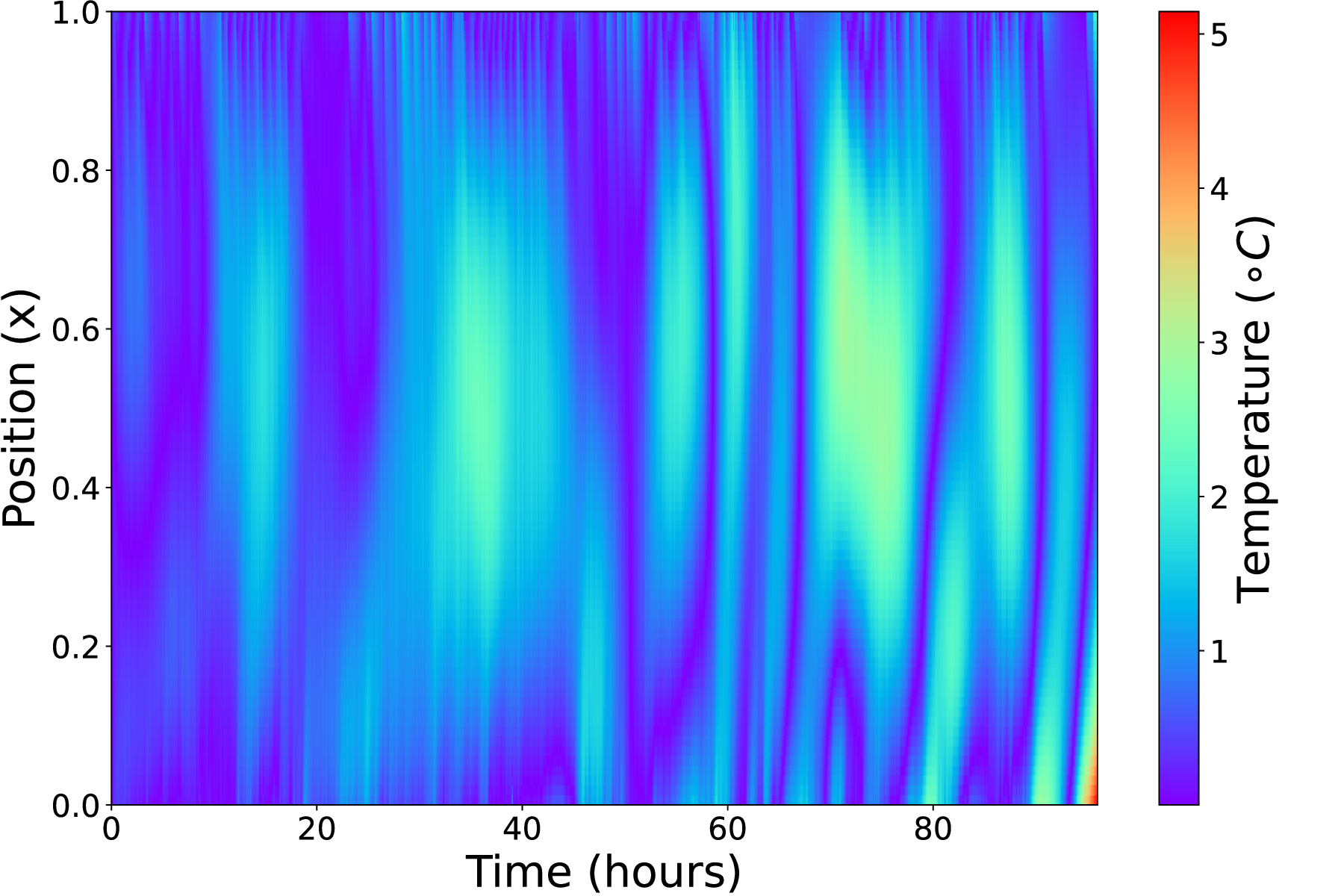}
	\includegraphics[width=\columnwidth]{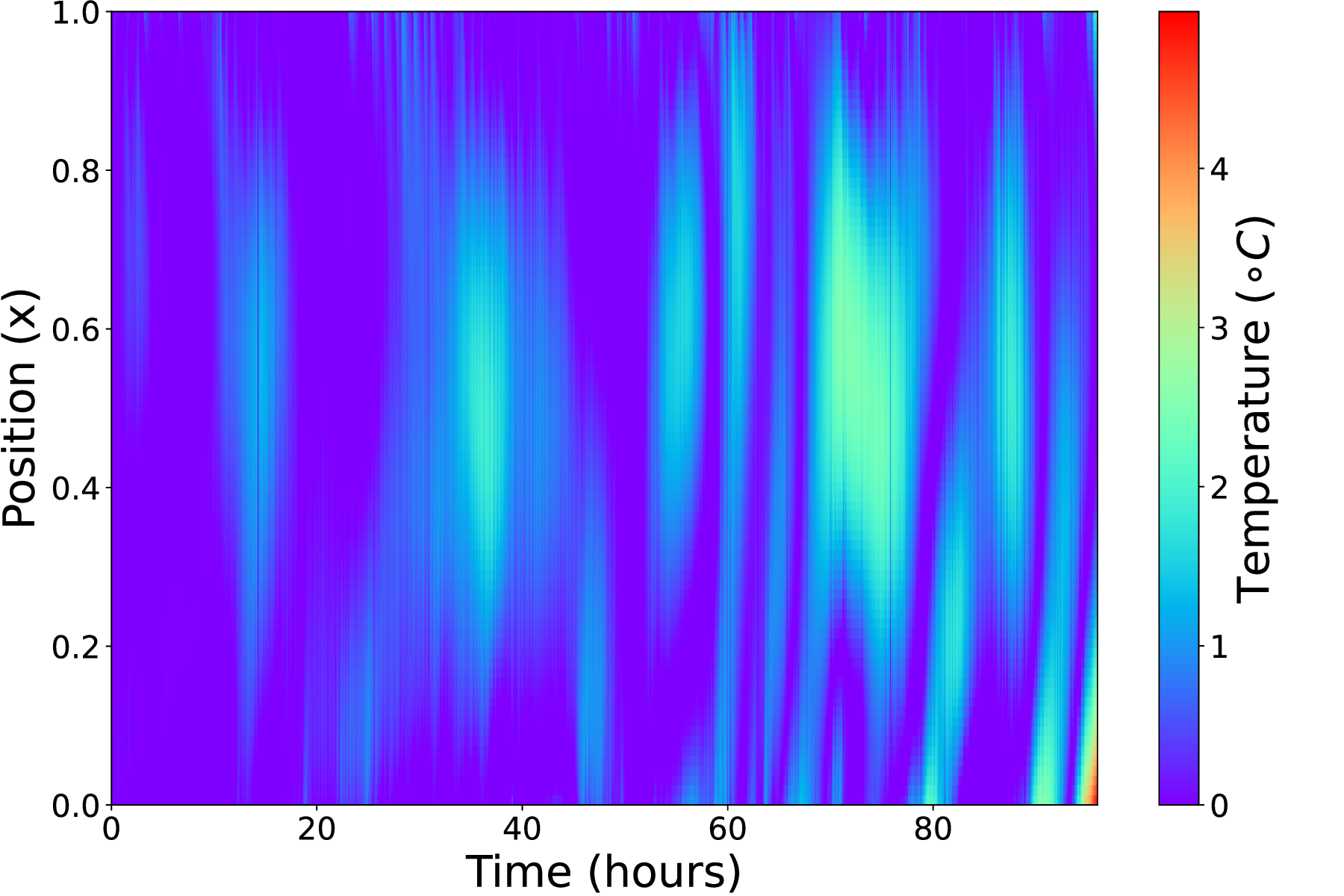}
	\caption{Error analysis of B-PINN predictions compared to FEM: (a) mean error; (b) uncertainty of error.}
	\label{fig:MeanTemperatureEstimation_BPINN_error}
\end{figure}

Comparisons with FEM reveal that the B-PINN not only reduces mean prediction error but also provides localized uncertainty estimates (Fig.~\ref{fig:MeanTemperatureEstimation_BPINN_error}). This enables operators to distinguish between confident and less reliable predictions.

\subsection{Discussion}
\label{SubSec:Discussion}

The B-PINN framework enhances PINNs by providing probabilistic, physics-consistent predictions. For transformer applications, this is valuable because:
\begin{itemize}
    \item Operators gain confidence intervals alongside predictions, improving decision-making.
    \item High-uncertainty regions highlight where additional sensing or simulation is most needed.
    \item Compared with deterministic PINNs, B-PINNs offer more robust generalization under scarce or noisy data.
\end{itemize}

Despite these benefits, B-PINNs remain computationally demanding due to sampling and variational inference overheads. Future research will explore more scalable Bayesian inference techniques and physics-informed priors to improve efficiency and robustness.

\section{Conclusions}
\label{Sec:Conclusions}

This two-part series has explored how neural network methodologies can be adapted and extended for transformer health monitoring. In Part~I, we reviewed the foundations of neural networks and their variants, including convolutional neural networks (CNNs) for acoustic monitoring and reinforcement learning (RL) for transformer energization control. These methods demonstrated how data-driven models can capture complex patterns and enable adaptive decision-making in transformer operation.

In this Part~II, we shifted focus toward the integration of physics and uncertainty into neural network learning. Physics-Informed Neural Networks (PINNs) were introduced as a framework that embeds governing physical laws directly into the training process. Applied to transformer thermal ageing, PINNs demonstrated the ability to generalize under scarce data, provide physically consistent predictions, and bridge data-driven learning with established thermal models. Building upon this, Bayesian PINNs (B-PINNs) extended the framework to account for epistemic uncertainty, yielding probabilistic spatiotemporal predictions and confidence intervals that are crucial for decision-making in asset management.

Taken together, these contributions outline a roadmap toward reliable, physics-aware, and uncertainty-informed machine learning for critical power assets. The following key insights emerge:
\begin{itemize}
    \item Data-driven methods (Part~I) excel in capturing patterns from rich sensor streams, but are limited when data is scarce or extrapolation is required.
    \item Physics-informed methods (Part~II) improve extrapolation and robustness by embedding domain knowledge, acting as implicit regularizers.
    \item Bayesian extensions provide uncertainty quantification, turning predictions into actionable insights by indicating confidence and highlighting where more sensing or modeling effort is needed.
\end{itemize}

Looking ahead, future research should focus on:
\begin{itemize}
    \item Developing scalable solvers for multi-physics and high-dimensional PINNs.
    \item Designing physics-informed priors and efficient inference methods to reduce the computational overhead of B-PINNs.
    \item Leveraging hybrid frameworks that integrate data-driven, physics-based, and probabilistic approaches within digital twins of transformers.
\end{itemize}

Overall, the fusion of physics, data, and uncertainty quantification opens the way for the next generation of trustworthy and efficient diagnostic and prognostic models, enabling more reliable and sustainable transformer fleet management.

\section*{Acknowledgements}

I am grateful to all my collaborators and mentors, especially Vic Catterson, Stephen McArthur, and Brian Stewart. I would also like to thank Xose M. Lopez-Fernandez for the invitation to present this tutorial at the ARWTr 2025 conference. In this Part II, I would like to acknowledge my collaborators on PINNs and B-PINNs: Ibai Ramirez, Jokin Alcibar, Joel Pino, Mikel Sanz, David Pardo, Luis del Rio, Alvaro Ortiz, and Kateryna Morozovska.

This work has been partially funded by the Spanish State Research Agency through the Ramón y Cajal Fellowship (grant No. RYC2022-037300-I) and Proyectos Generación de Conocimiento (grant No. PID2024-156284OA-I00), co-funded by MCIU/AEI/10.13039/501100011033 and FSE+.

\bibliographystyle{IEEEtran}

\bibliography{IEEEabrv,myBib}

\end{document}